%% file: 00root.tex
\documentclass[letterpaper, 10pt, conference]{ieeeconf}
\IEEEoverridecommandlockouts
\usepackage{cite}
\usepackage{amsmath,amssymb,amsfonts}
\usepackage{algorithm}
\usepackage{multirow}
\usepackage{algpseudocode}
\usepackage{graphicx}
\usepackage{textcomp}
\usepackage{hyperref}
\usepackage{xcolor}
\usepackage{booktabs}

\usepackage{xcolor}

\newcommand{\model}[0]{USA-Net}

\newcommand{\method}[0]{USA-Net}

\newcommand{\beginsupplement}{%
        \setcounter{table}{0}
        \renewcommand{\thetable}{S\arabic{table}}%
        \setcounter{figure}{0}
        \renewcommand{\thefigure}{S\arabic{figure}}%
     }

\def\BibTeX{{\rm B\kern-.05em{\sc i\kern-.025em b}\kern-.08em
    T\kern-.1667em\lower.7ex\hbox{E}\kern-.125emX}}
\begin{document}

\title{\LARGE \bf
USA-Net: Unified Semantic and Affordance Representations \\
for Robot Memory
}


\author{
Ben Bolte, Austin Wang, Jimmy Yang, Mustafa Mukadam, Mrinal Kalakrishnan, Chris Paxton
\thanks{All authors are with Meta AI.}
\thanks{Address correspondence to \texttt{bbolte@meta.com}}
}

\maketitle

\begin{abstract}
\input{01abstract.tex}
\end{abstract}

\section{INTRODUCTION}
\input{02intro}

\section{Related Work}
\input{03related}

\section{Methodology}
\input{04methodology}

\section{Experiments}
\input{05experiments}

\section{Conclusion}
\input{06conclusion}

\bibliographystyle{plain}
\bibliography{refs}

\beginsupplement

\end{document}

%% file: 01abstract.tex
In order for robots to follow open-ended instructions like ``\textit{go open the brown cabinet over the sink},'' they require an understanding of both the scene geometry and the semantics of their environment.
Robotic systems often handle these through separate pipelines, sometimes using very different representation spaces, which can be suboptimal when the two objectives conflict.
In this work, we present \method{}, a simple method for constructing a world representation that encodes both the semantics and spatial affordances of a scene in a differentiable map.
This allows us to build a gradient-based planner which can navigate to locations in the scene specified using open-ended vocabulary.
We use this planner to consistently generate trajectories which are both shorter 5-10\% shorter and 10-30\% closer to our goal query in CLIP embedding space than paths from comparable grid-based planners which don't leverage gradient information.
To our knowledge, this is the first end-to-end differentiable planner optimizes for both semantics and affordance in a single implicit map.
%
%
Code and visuals are available at our website: \href{https://usa.bolte.cc/}{usa.bolte.cc}

%% file: 02intro.tex
%
In order for a robot to complete a request such as, ``\textit{take this can to the kitchen}'', it needs \textit{semantic} information about the environment, such as the location of the kitchen, as well as \textit{affordance} information which will allow it to develop a motion plan that does not collide with other objects in the scene.
In this work, we propose \textit{\textbf{U}nified \textbf{S}emantic and \textbf{A}ffordance Representations for Robot Memory}~(\model{}), which combines two approaches for building differentiable maps for encoding the semantics and affordance of a scene, allowing us to learn both in a single network.
We use this network to implement a differentiable planner that jointly optimizes for both finding a collision-free trajectory and achieving a semantic goal.

%
The core problem we consider is building a world representation which enables end-to-end differentiable planning for semantic goals in cluttered environments.
In order to build this representation, prior work commonly uses off-the-shelf image classification models trained on web-scale datasets~\cite{clip-fields,openscene,nlmap-saycan,jatavallabhula2023conceptfusion}, especially CLIP~\cite{clip} and Detic~\cite{detic}.
%
If the system is able to accurately estimate the robot's pose and the depth of objects in the scene, then the semantics of the image in camera space can be projected to provide each surface point in the scene with a semantic label.
This semantic information can then be used to find the location of objects for planning~\cite{clip-fields,jatavallabhula2023conceptfusion}.
%

\begin{figure}[bt]
\centering
\includegraphics[width=\linewidth]{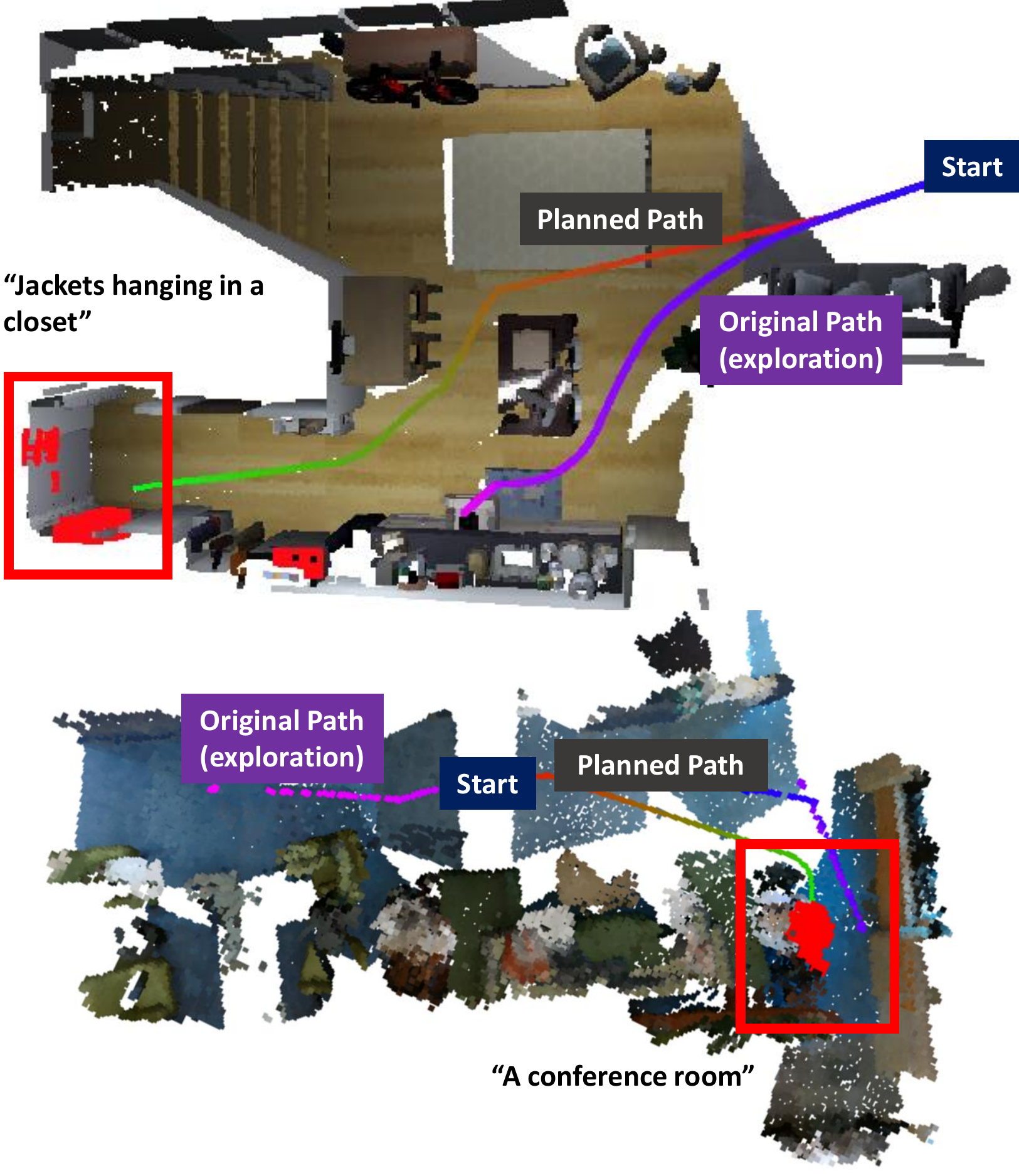}
\caption{\method{} lets us jointly optimize for satisfying language goals, and finding collision-free trajectories. Above, see an example trajectory from a random starting point sampled from the camera trajectory, navigating to two different target queries, ``\textit{Jackets hanging in the closet}'' and ``\textit{a conference room}.'' The camera poses are shown in pink and blue. The generated trajectory is shown starting in red and ending in green. High semantic similarity points are shown in bright red. For more examples, see our website.}
\label{fig:point_cloud}
\vskip -0.5cm
\end{figure}

However, this leaves the question of what trajectory should be taken to reach the goal without colliding with obstacles.
Often, for navigating to goals, we can use discrete or sampling-based planners to navigate around detected obstacles~\cite{gervet2022navigating}. 
However, these assume that we know the exact goal location to move to; in the case of a learned representation, we instead see a softer distribution of \textit{potential} goals, where there may be tradeoff between trajectory planning and goal selection.
For example, the optimal semantic location may be blocked by an obstacle or otherwise unreachable.
Additionally, course-grained discrete motion planners, such as occupancy grids which use large cell sizes to conserve memory, may fail to capture fine-grained semantic information, such as small details of a scene.
For example, to execute the command, ``\textit{grab a soda from the refrigerator}'', a robot must be able to navigate around course-grained obstacles in the environment, but also interact with the relatively fine-grained soda can.

To handle such cases, we look to recent work on building implicit, differentiable representations for robot motion planning like iSDF~\cite{isdf}. Implicit representations can be used to capture distances from scene geometry, which in turn can be used for motion planning. CLIP-Fields~\cite{clip-fields} also showed that these representation can capture the semantic information necessary for open-vocabulary object navigation, but did not look at how to use these for motion planning. This poses an obvious question: how can we use RGB and depth data to capture a representation which can be used for both semantic goal navigation and motion planning?


To this end, we combine the differentiable affordance representation captured by iSDF~\cite{isdf} with the semantic representation based on CLIP-Fields~\cite{clip-fields}. We find that we are able to use this approach to jointly supervise the semantic representation of empty space, by computing the CLIP embedding of the batchwise nearest neighbor point in addition to using it to compute the ground truth SDF value. This allows us to build a representation containing gradient information for both semantics and affordances. We then present a novel gradient-based planning algorithm which operates on this implicit representation, and optimizing both the semantic goal and trajectory constraints at once, achieving stronger results than we could by optimizing only one of them.


The paper’s contributions are three-fold:
\begin{enumerate}
    \item We propose \model{}, an implicit representation which captures both semantic and affordance information, and can be used to build a high-quality motion planner for navigating to  open-ended goals specified by natural language.
    \item We develop supervision mechanisms which leverage the supervision signal from the affordance planner to jointly supervise the semantics of empty space, giving us a semantic map which is differentiable throughout the entire scene.
    \item Finally, we evaluate~\model{}'s gradient-based planner which achieves up to 91.2\% and 39.2\% higher semantic similarity and shorter trajectories, respectively, over prior discrete planners.
\end{enumerate}


%% file: 03related.tex
\subsection{Vision-language navigation}

In this work, we consider building a semantic-affordance map for a novel environment, in order to have a robot navigate to open-set natural language goals.
Recent work has applied natural language to robotic navigation in simulated environments~\cite{janner2018representation,misra2017mapping} or 3D realistic environments~\cite{anderson2018vision} such as Doom~\cite{kempka2016vizdoom}, AI2-THOR~\cite{shridhar2020alfred,kolve2017ai2}, and House3D~\cite{wu2018building}.
However, much of this work leaves unaddressed how to build ideal representations for planning, and focuses on high-level or discrete actions~\cite{shridhar2020alfred}. While many solutions for complex vision-language navigation tasks build either a 2D or 3D map~\cite{min2021film,gervet2022navigating,blukis2022persistent}, these approaches build their maps \textit{explicitly}, specifically updating obstacle locations and object classes.
Recently, a number of works have begun using CLIP features to build spatial representations with a deep semantic understanding~\cite{gadre2022clip,huang2022visual,chen2022open,clip-fields}. CLIP-on-wheels~\cite{gadre2022clip} showed CLIP~\cite{clip} could be used to find uncommon objects.
To build a system that can operate in the real world, NLMap-SayCan~\cite{chen2022open} and~VLMap~\cite{huang2022visual} use CLIP features~\cite{clip} to build a semantic representation of the environment, and then employ a large language model to produce a sequence of sub-instructions grounded in the pre-built semantic map.
ConceptFusion~\cite{jatavallabhula2023conceptfusion} likewise builds a 3D multi-modal map which can be used for many different types of queries.
However, while these maps capture geometry, they are either compute-expensive to build, or rely on interfacing with complex downstream systems to translate high-level semantic targets into low-level motion plans.
In contrast, \method{} is a unified framework that jointly learns the representation that encodes both semantic and avoidance information.

\subsection{Motion planning}

When doing motion planning in an unmodeled environment, we often use an occupancy map to represent a discrete approximation of the continuous space.
Much prior work on real-world object navigation~\cite{gervet2022navigating,vlmaps} has used this sort of formulation.

%
%
Several algorithms can be used to find an optimal path in the occupancy map, such as A* search algorithm~\cite{a-star}, RRT*~\cite{rrt-star}, fast-matching trees (FMT*)~\cite{fmt}, batch informed trees (BIT*)~\cite{bit}, adaptively informed trees (AIT*)~\cite{ait}, effort informed trees (EIT*)~\cite{eit}, and many others.
Methods like MP-Nets~\cite{qureshi2020motion} have been able to combine these sorts of sampling-based planners with learned models for faster, optimal search.

While these algorithms provide assorted guarantees about either the search for the path or its optimality, they are often constructed independently from maps which encode the scene semantics, which can result in suboptimal performance in cases where there are tradeoffs between the path produced by the motion planner and the provided semantic goal.
Planners like CHOMP~\cite{ratliff2009chomp} use gradient-based optimization to improve upon some trajectory cost function, which can then be used to create an optimal trajectory by balancing multiple objectives. Recently, Motion Policy Nets~\cite{fishman2022motion} applied a similar technique to generating optimal trajectories in fixed base manipulation problems.

None of these planners, however, provide an obvious interface to natural-language goals, except by predicting the goal~\cite{vlmaps,nlmap-saycan}.
Various methods have been proposed to add language based goals and constraints into the motion planning problem. Sharma et al.~\cite{sharma2022correcting} proposed a method which combined a model predictive control (MPC) planner~\cite{bhardwaj2022storm} with learned cost functions. Also closely related is work which learns to predict reward functions from natural language instructions~\cite{williams2018learning}, although we focus specifically on optimal motion planning and open-vocabulary language.

%% file: 04methodology.tex
\begin{figure}[tb]
    \includegraphics[width=\linewidth]{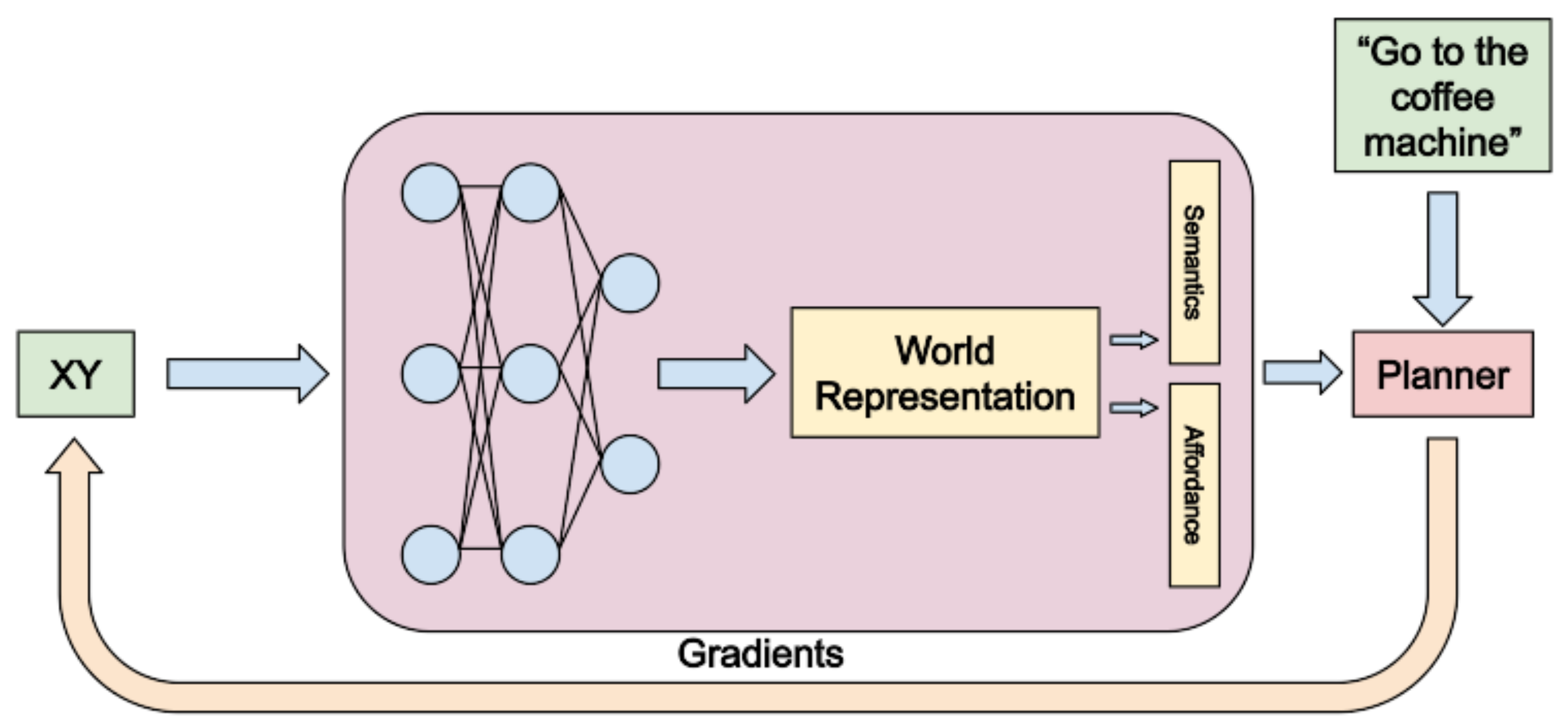}
    \caption{A high-level diagram of the system. A single network encodes both the semantic and affordance information for the environment, enabling the robot to navigate using open-ended vocabulary queries.}
    \label{fig}
\end{figure}

Our approach to building our representation takes as input a sequence of RGB images with corresponding per-pixel depths, as well as the camera pose and intrinsics at each frame. These can be estimated online \cite{hector-slam} or offline \cite{colmap}. These inputs are common in robotics applications and can be used to construct point clouds of the scene. In our real-world system, we found that using the depth and RGB images from an Intel RealSense camera~\cite{intel-realsense} and the poses from Hector SLAM~\cite{hector-slam}, despite being noisier than offline methods, were sufficient for our application.

In order to construct a single representation which models both semantic and affordance, we sample frames from a replay buffer, and sample $X, Y, Z$ points from the frustrum of each frame. We then pass the sampled coordinates to our representation network to give a feature vector for each point in space. We train this feature vector to encode both the semantic and affordance information of the physical point.

\subsection{Affordance Representation}

To supervise the affordance prediction for our world representation, we train our model to predict the Signed Distance Function (SDF) as in \cite{isdf}. To obtain a ground-truth value, we experimented with first building a point cloud for the entire scene and querying the global nearest neighbor point, as well as with using the batchwise sampling approach described in the original work. We found that the latter approach provided estimates of the SDF which closely matched the globally-computed SDF, while being simpler to construct. We use the mean squared error loss function

\begin{equation}
    \label{eq:affordance-loss}
    \mathcal{L}_r = (\hat{r} - \delta_{SDF})^2
\end{equation}

\noindent where:

\begin{itemize}
    \item $\hat{r}$ is the affordance representation vector from our model
    \item $\delta_{SDF}$ is the ground truth signed distance value for each point
\end{itemize}

These approaches worked well out-of-the-box in cases where we were able to obtain very accurate depth estimates such as simulated data from ReplicaCAD~\cite{szot2021habitat}. However, in cases with noisy depth estimates -- such as those from the Intel RealSense depth camera -- the global SDF estimate appeared to significantly underestimated the true SDF value, resulting in predicted SDFs which were too small when compared with the occupancy grid generated from the raw point cloud, as shown in Figure \ref{fig:noise_correction}.
This made our final planner overly cautious. The effect was much larger than the expected effect of \textit{over}estimating the true SDF value described in~\cite{isdf}. This resulted in affordance representations which were significantly degraded when compared with a baseline occupancy grid.

For this reason, we found that using the batchwise sampling approach greatly improved our estimated SDF. However, when this model was trained on data captured with our robot, we found that it \textit{still} systematically underestimated the true SDF, making our planner overly cautious around obstacles compared to our baseline planner. We therefore found it important to attempt to model this bias, so that we could correct for it in our final planner.

\subsubsection*{Estimating the SDF Bias}

We model the depth $D_{ij}$ for pixel $p_{ij}$ using a normal distribution with standard deviation $\sigma_D$ centered at the true depth $d_{ij}$:

$$D_{ij} \sim \mathcal{N}(d_{ij}, \sigma_D^2)$$

\begin{figure}[bt]
    \includegraphics[width=\linewidth]{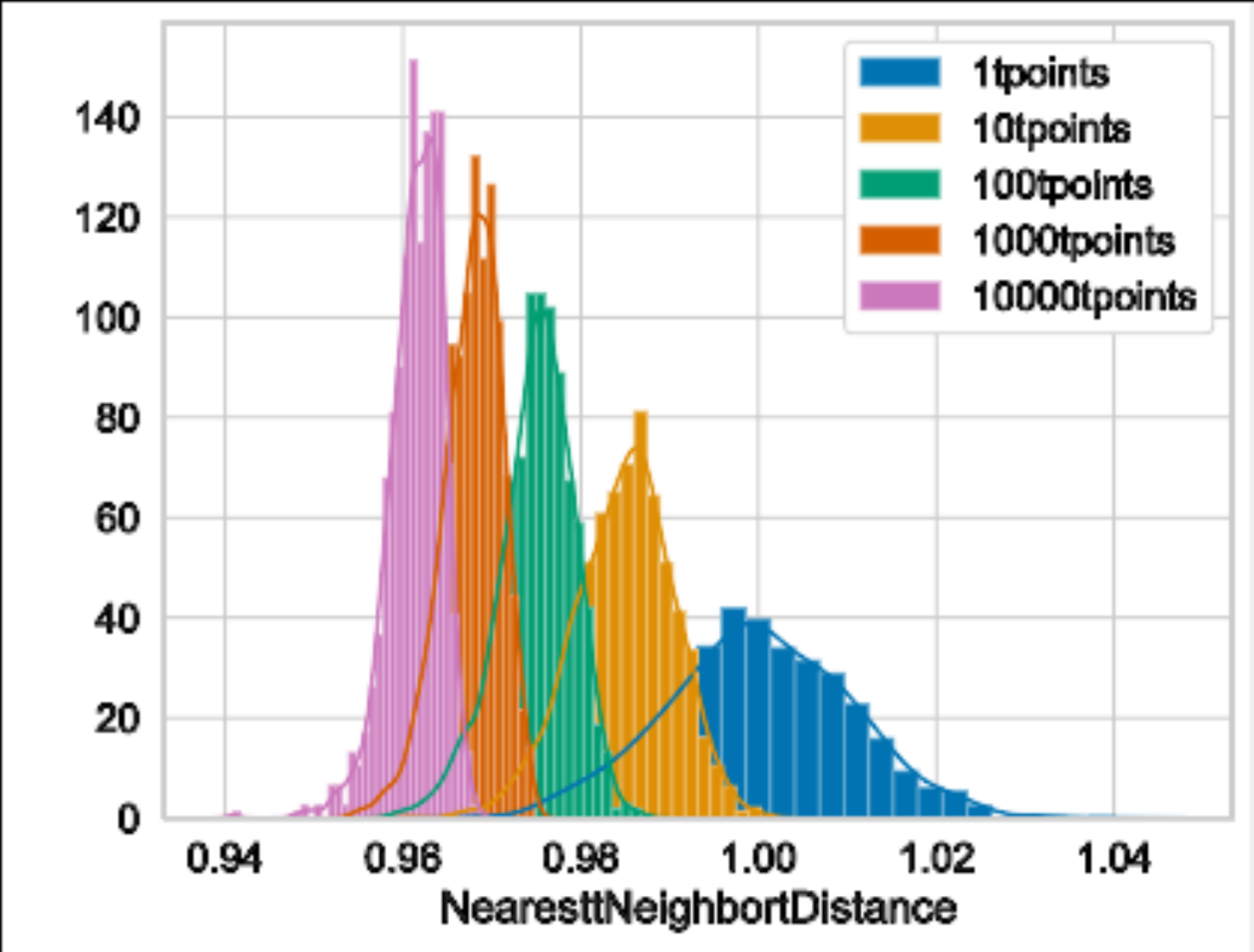}
    \caption{Simulation showing how sampling more points systematically increases the bias in the final SDF estimate. In this case, we consider the SDF computed using the nearest neighbor point sampled from $N$ points centered at the same location 1 meter from the reference point (i.e., a true SDF value of 1 meter), drawn from the distribution described in Equation \ref{eq:point_distribution} with $\sigma_C = 1 cm$.}
    \label{fig:distribution_simulation}
\end{figure}

\begin{figure}[bt]
    \centerline{\includegraphics[width=0.8\linewidth]{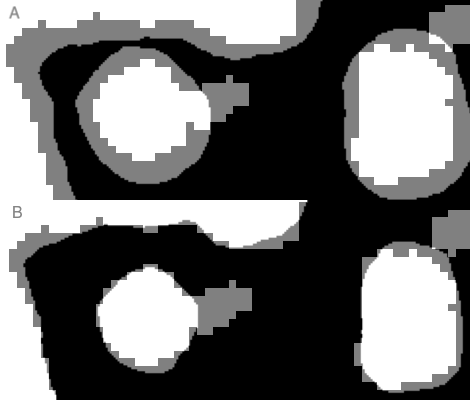}}
    \caption{An illustration of the underestimation modeled by Equation \ref{eq:point_distribution}. In the top image, we show the navigable space found by thresholding the SDF to 50 centimeters, overlaid on navigable space found by taking the Minkowski sum of the point cloud-based occupancy grid with a 50 centimeter diameter column. The SDF underestimates the navigable space relative to the point cloud-based occupancy grid. On the bottom, we instead threshold the SDF to 30 centimeters, estimated as in Figure~\ref{fig:distribution_simulation}, and find that it aligns much more closely with the point cloud-based navigable space.}
    \label{fig:noise_correction}
\end{figure}

Projecting the pixel $p_{ij}$ using the depth distribution from the camera's frame to the point cloud coordinate $\hat{c}_{ij}$ adds additional error due to noise in the camera's pose. This can be modeled using another normal distribution with standard deviation $\sigma_P$. Since this transformation is additive, we can model the location of each point in the point cloud as:

\begin{equation}
    \label{eq:point_distribution}
    \begin{aligned}
    C_{ij} & \sim \mathcal{N}(\hat{c}_{ij}, \sigma_D^2 + \sigma_P^2) \\
    & \sim \mathcal{N}(\hat{c}_{ij}, \sigma_C^2) \\
    \end{aligned}
\end{equation}

Now, consider the case where we aggregate $N$ points at the same location, such as would happen for a stationary robot, or for points which are very close to each other in the map frame relative to the variance of the depth or pose distributions. If we use this point when computing the SDF, we would actually choose the closest point from all of these aggregated points, which is the source of the bias in our final ground truth SDF. This is illustrated in Figure \ref{fig:distribution_simulation}. We use realistic estimates for the noise in our depth camera, pose and point density to obtain a constant adjustment to our final SDF of 10 to 20 centimeters in our real-world robot.

\subsubsection*{Baseline Affordance Representations}

We compare our model's affordance representation with an occupancy map constructed by projecting points to their $\left(X, Y, Z\right)$ coordinates and determining if any of the points in a given grid cell would obstruct our robot. To account for noise in this representation, we take the moving average of each grid cell in the occupancy map over each frame in the clip, with 0 for unoccupied and 1 for occupied. To get our final binary occupancy map, we simply threshold of this value. We found that this approach was relatively robust to different constants for thresholding and moving average update.

While the occupancy map is simple to construct and update over time, it suffers from the tradeoff between decreasing the size of each grid cell and increasing the computational demands on both the mapper and planner. We chose a grid cell size of 10 centimeters when mapping our environments based on the size of our robot, although this will likely will vary for different agents.

\subsection{Semantic Representation}

In order to obtain a semantic representation for each physical point in space, we use a pre-trained CLIP model~\cite{clip}. We adapt a modified CLIP-Fields~\cite{clip-fields} which allows us to supervise any given $\left(X, Y, Z\right)$ coordinate in the same 3D space as our affordance representation.

In contrast with many approaches for building semantic representations of scenes, we wanted our representation to be continuous, so that our downstream planner could jointly optimize the target location along with the path. However, this presents the problem of determining the semantic representation of empty space.

Our solution to this issue is to supervise the semantic representation and affordance representation using the same nearest neighbor point. Since the nearest surface point to the coordinate in space is likely to have a well-defined semantic meaning (for example, "wall" or "chair"), we supervise the semantic meaning of the point in empty space using the CLIP embedding for that surface point. We supervise our representation using the same contrastive loss used to train the original CLIP model, which preserves the meaningfulness of the dot product of the learned embedding vector with the CLIP embedding of arbitrary linguistic queries. Additionally, we found that weighting the per-sample loss by the SDF helped improve the model's convergence, by penalizing points less if they are further away from the surface which was used to compute the CLIP embedding. This makes intuitive sense, since points which are further away from some surface points have a less well-defined semantic meaning.

Thus, we supervise our semantic representation using the loss function:
\begin{equation}
    \label{eq:semantic-loss}
    \mathcal{L}_{s} = \text{softmax}(\delta_{SDF}) \text{softmax}(\hat{s} \cdot s)
\end{equation}

\noindent where:

\begin{itemize}
    \item $\hat{s}$ is the semantic representation vector from our model
    \item $s$ is the ground truth CLIP embedding
    \item $\delta_{SDF}$ is the ground truth signed distance value for each point
    \item $\text{softmax}(\hat{s} \cdot s)$ is the normal CLIP loss
\end{itemize}

Combining equations \ref{eq:affordance-loss} and \ref{eq:semantic-loss} gives us our final loss function for supervising the network:

\begin{equation}
    \label{eq:final-loss}
    \mathcal{L} = \lambda_r \mathcal{L}_{r} + \lambda_s \mathcal{L}_{s}
\end{equation}

where $\lambda_r$ and $\lambda_s$ are weighting constants. We found that the choice of loss weight was not particularly important and weight the two terms equally.

\begin{algorithm}[bt]
\caption{\textit{Grid planner}: Planner that discretizes the affordance values to build an occupancy grid, then performs a grid-based search such as $A \ast$ or $RTT$ to compute the optimal path.}
\label{alg:grid}

\hspace*{\algorithmicindent} \textbf{Input}
\begin{description}
    \item Representation function $r$
    \item Grid size $s$
    \item Minimum distance to obstacles $d_{min}$
    \item Initial position $p_0$
    \item Target location query $q$
\end{description}

\hspace*{\algorithmicindent} \textbf{Output}
\begin{description}
    \item Way points $\hat{p}$
\end{description}

\begin{algorithmic}[1]
\State Initialize an empty grid $G = \{\}$
\State Initialize the maximum CLIP score $s_{max} = -\infty$
\For {$x, y \in \{x_{min}, ...., x_{max}\}, \{y_{min}, ..., y_{max}\}$}
    \State Query $r(x, y)$ to get $sdf$ and $clip$
    \If {$sdf > d_{min}$}
        \State Add $[x, y]$ to $G$
        \If {$clip > s_{max}$}
            \State Update $p_T = [x, y]$ and $s_{max} = clip$
        \EndIf
    \EndIf
\EndFor
\State \Return $\hat{p} = Search(p_0, p_T, G)$  \Comment{$A \ast$ or similar}
\end{algorithmic}

\end{algorithm}

\subsection{Grid Motion Planner}

The main point of building a shared representation space is to simplify the construction of the motion planner. At a high level, the planner takes the current location and a semantic query representing the goal location, and outputs a path which the robot can use to navigate to the target location. We describe two approaches for solving this task, which we call \emph{grid} (Algorithm \ref{alg:grid}) and \emph{gradient} (Algorithm \ref{alg:gradient}) planners.

\subsubsection*{Baseline Motion Planner}

To establish a baseline planner to compare our affordance representation against, we use an occupancy map generated from the raw point cloud. We then generate way points using a grid-based planner. In order to establish a strong baseline, we use breadth-first search (BFS) to find the shortest path between our starting and ending points. Additionally, we remove intermediate points which are within line-of-sight of each other as determined using Bresenham's line algorithm. 

\subsubsection*{Occupancy Grid from World Representation}

In the \emph{grid} motion planner, we simply convert the world representation to a grid, then perform a grid-based search to compute way points, as in the baseline planner. We consider a grid cell to be "occupied" if any corner has an SDF value smaller than the radius of our robot plus some buffer. When given a target query rather than a specific goal location, we choose the grid cell with the highest similarity to the query vector as our target location.

This approach has a couple of advantages over the baseline planner. For one, we can easily and quickly generate a new grid with different cell sizes, since the affordance function is represented by our network, whereas in our baseline planner doing so would require keeping the entire point cloud in memory. Additionally, we found that the occupancy grids generated from our world representation were smoother than grids generated from the point cloud, since the network acts as a regularizer.

\begin{algorithm}[bt]
\caption{\textit{Gradient planner}: End-to-end planner which optimizes the way points to maximize an objective function.}
\label{alg:gradient}

\hspace*{\algorithmicindent} \textbf{Input}
\begin{description}
    \item Representation function $r$
    \item Number of way points $N_p$
    \item Number of initial target points to sample $N_t$
    \item Minimum distance to obstacles $d_{min}$
    \item Initial position $p_0$
    \item Objective weights $\lambda_o$, $\lambda_n$, $\lambda_s$, $\lambda_d$
    \item Target location query $q$
\end{description}

\hspace*{\algorithmicindent} \textbf{Output}
\begin{description}
    \item Way points $\hat{p}$
\end{description}

\begin{algorithmic}[1]
\State Sample target point $p_T = \max_{i} clip(x_i, y_i) \cdot q$
\State Initialize $\hat{p}$
\While {not converged}
    \State Compute $\mathcal{L}_o$ \Comment{See Equation \ref{eq:loss_o}}
    \State Compute $\mathcal{L}_n$ \Comment{See Equation \ref{eq:loss_n}}
    \State Compute $\mathcal{L}_s$ \Comment{See Equation \ref{eq:loss_s}}
    \State Compute $\mathcal{L}_d$ \Comment{See Equation \ref{eq:loss_d}}
    \State $\mathcal{L} = \lambda_o \mathcal{L}_o + \lambda_n \mathcal{L}_n + \lambda_s \mathcal{L}_s + \lambda_d \mathcal{L}_d$
    \State Update $\hat{p}$ to minimize $\mathcal{L}$
\EndWhile

\State \Return $\hat{p}$
\end{algorithmic}

\end{algorithm}

\subsection{Gradient Motion Planner}

In the \emph{gradient} motion planner, we use the differentiable nature of the world representation to consider the following objectives:

\begin{itemize}
    \item Do not run into obstacles
    \item Keep the way points roughly evenly spaced
    \item Move the final way point close to the target query
    \item Make the path shorter
\end{itemize}

Formally, we define differentiable loss functions for each objective and optimize the way points using gradient descent. The flexible nature of the world representation makes it easy to optimize any objective; we chose these ones because they seem to produce reasonable paths in practice.

\subsubsection*{Do not run into obstacles}

We define the loss function

\begin{equation}
    \label{eq:loss_o}
    \mathcal{L}_o = \sum_{p_i \in \hat{p}} \text{clamp}(d_{min} - sdf(p_i), 0, \infty)
\end{equation}

where

\begin{itemize}
    \item $\hat{p}$ are the way points in the current path
    \item $d_{min}$ is the minimum distance to an obstacle
    \item $sdf(p_i)$ is the SDF returned by our world representation
\end{itemize}

We clamp this function to avoid penalizing points which are already sufficiently far away from any nearby objects.

\subsubsection*{Keep the way points roughly evenly spaced}

We define the loss function

\begin{equation}
    \label{eq:loss_n}
    \mathcal{L}_n = \sum_{p_i \in \hat{p}} \text{abs}(\text{norm}(p_i, p_{i + 1}) - \text{norm}(p_i, p_{i - 1}))
\end{equation}

This encourages points to be roughly evenly spaced between their left and right neighbors. Without adding this loss function, it is possible for two adjacent way points to be on either side of an obstacle, resulting in an invalid path. Keeping the way points close together minimizes the risk of this happening. This component of the loss does not require us to query the network.

\subsubsection*{Move the final way point close to the target query}

We define the loss function

\begin{equation}
    \label{eq:loss_s}
    \mathcal{L}_s = -(clip(p_T) \cdot q)
\end{equation}

\noindent where:

\begin{itemize}
    \item $clip(p_T)$ is the CLIP embedding returned from our world representation for the terminal point $p_T$
    \item $q$ is the CLIP embedding for the target query
\end{itemize}

This loss function encourages the point to move in the direction which maximizes the CLIP score for the final point.

\subsubsection*{Make the path shorter}

We define the loss function

\begin{equation}
    \label{eq:loss_d}
    \mathcal{L}_d = \sum_{p_i \in \hat{p}} \text{norm}(p_{i}, p_{i+1})
\end{equation}

This is simply the total path length. Like Equation \ref{eq:loss_n}, this component of the loss does not require us to query the network.

\subsubsection*{Choosing the Initial Path}

Since the gradient planner is simply doing gradient descent to optimize the chosen objective, a poor initial path can greatly reduce the performance or prevent it from converging. Choosing a good initial path is therefore very important. A simple initial path can be a straight line between the starting and (initial) ending points. Alternatively, the path can be initialized using the output of another planner (such as the grid planner). We used the latter approach throughout our experiments. This is important for finding a good-quality target location for a given semantic query, because we found that the gradient information in the semantic map was only locally useful; if the initial goal location was too far away from a good goal location, then our gradient planner would fail to converge.

\subsubsection*{Conflicting Loss Objectives}

There are some cases when the loss objectives conflict. For example, if the optimal semantic location is not reachable, then the directions of the gradients to minimize the losses in equations \ref{eq:loss_o} and \ref{eq:loss_d} will be in opposite directions. We therefore weight each loss term to get our final loss using the weights $\lambda_o = 1$, $\lambda_n = 1$, $\lambda_s = 15$ and $\lambda_d = 25$. We experimented with different optimizers and found that Adam with a learning rate of $1e-2$ works reasonably well.

%% file: 05experiments.tex
In this section, we quantitatively evaluate the performance of our motion planner in a variety of scenes recorded with our physical robot. In our evaluation, we build representations for three heterogeneous scenes recorded with the Stretch from Hello Robot~\cite{stretch-robot}; one from our lab (the \textit{lab} scene), one from a nearby kitchen (the \textit{kitchen} scene), and one of a pair of chess boards (the \textit{chess} scene). We additionally evaluate on two scenes recorded using the Record3D iPhone application, and one scene recorded from the ReplicaCAD simulation~\cite{szot2021habitat}. A complete collection of the generated trajectories are available on the website.

\begin{table}[bt]
\centering
\caption{Average length multiple over the shortest possible path between 100 randomly chosen starting and ending point pairs. A lower value indicates a shorter average path length. A value of $1$ indicates that the path lengths for a given planner are monotonically better than the path lengths from all other planners for a given scene. For a visualization of the paths and environments, see the website associated with this paper.}
\label{tab:path_lengths}

\begin{tabular}{llc}
\toprule
\multirow{4}{*}{Lab (Robot)}  & $10cm$ Baseline        & 1.029 \\
                & $10cm$ Grid Planner         & 1.022 \\
                & $20cm$ Grid Planner         & 1.032 \\
                & $30cm$ Grid Planner         & 1.050 \\
                & $40cm$ Grid Planner         & 1.151 \\
                & Gradient Planner            & \textbf{1.000} \\ \hline
\multirow{4}{*}{Kitchen (Robot)}  & $10cm$ Baseline        & 1.422 \\
                & $10cm$ Grid Planner         & 1.026 \\
                & $20cm$ Grid Planner         & 1.116 \\
                & $30cm$ Grid Planner         & 1.134 \\
                & $40cm$ Grid Planner         & 1.563 \\
                & Gradient Planner            & \textbf{1.000} \\\hline
\multirow{4}{*}{Chess (Robot)}  & $10cm$ Baseline        & 1.068 \\
                & $10cm$ Grid Planner         & 1.005 \\
                & $20cm$ Grid Planner         & 1.010 \\
                & $30cm$ Grid Planner         & 1.010 \\
                & $40cm$ Grid Planner         & 1.065 \\
                & Gradient Planner            & \textbf{1.000} \\\hline
\multirow{4}{*}{Lab (iPhone)}  & $10cm$ Baseline        & \textbf{1.000} \\
                & $10cm$ Grid Planner         & 1.045 \\
                & $20cm$ Grid Planner         & 1.067 \\
                & $30cm$ Grid Planner         & 1.098 \\
                & $40cm$ Grid Planner         & 1.124 \\
                & Gradient Planner            & 1.024 \\\hline
\multirow{4}{*}{Studio (iPhone)}  & $10cm$ Baseline        & 1.003 \\
                & $10cm$ Grid Planner         & 1.035 \\
                & $20cm$ Grid Planner         & 1.075 \\
                & $30cm$ Grid Planner         & 1.157 \\
                & $40cm$ Grid Planner         & 1.127 \\
                & Gradient Planner            & \textbf{1.000} \\\hline
\multirow{4}{*}{ReplicaCAD~\cite{szot2021habitat}}  & $10cm$ Baseline        & 1.048 \\
                & $10cm$ Grid Planner         & 1.006 \\
                & $20cm$ Grid Planner         & 1.013 \\
                & $30cm$ Grid Planner         & 1.022 \\
                & $40cm$ Grid Planner         & 1.024 \\
                & Gradient Planner            & \textbf{1.000} \\
\bottomrule
\end{tabular}
\end{table}

\begin{table}[bt]
\centering
\caption{Average relative clip scores for the final location for a set of semantic queries for various scenes. A higher score indicates that the final position is closer in CLIP embedding space to the target query. A value of $1$ indicates that the planner was monotonically better than all other planners for each of the test queries in a given scene. To visualize the planned paths, see the associated website.}
\label{tab:semantics}

\begin{tabular}{llc}
\toprule
\multirow{4}{*}{Lab (Robot)}  & $10cm$ Grid Planner         & 0.955 \\
                & $20cm$ Grid Planner         & 0.910 \\
                & $30cm$ Grid Planner         & 0.822 \\
                & $40cm$ Grid Planner         & 0.829 \\
                & Gradient Planner            & \textbf{0.999} \\ \hline
\multirow{4}{*}{Kitchen (Robot)}  & $10cm$ Grid Planner         & 0.979 \\
                & $20cm$ Grid Planner         & 0.889 \\
                & $30cm$ Grid Planner         & 0.853 \\
                & $40cm$ Grid Planner         & 0.775 \\
                & Gradient Planner            & \textbf{1.000} \\\hline
\multirow{4}{*}{Chess (Robot)}  & $10cm$ Grid Planner         & 0.936 \\
                & $20cm$ Grid Planner         & 0.886 \\
                & $30cm$ Grid Planner         & 0.768 \\
                & $40cm$ Grid Planner         & 0.634 \\
                & Gradient Planner            & \textbf{1.000} \\\hline
\multirow{4}{*}{Lab (iPhone)}  & $10cm$ Grid Planner         & 0.978 \\
                & $20cm$ Grid Planner         & 0.947 \\
                & $30cm$ Grid Planner         & 0.867 \\
                & $40cm$ Grid Planner         & 0.752 \\
                & Gradient Planner            & \textbf{1.000} \\\hline
\multirow{4}{*}{Studio (iPhone)}  & $10cm$ Grid Planner         & 0.985 \\
                & $20cm$ Grid Planner         & 0.923 \\
                & $30cm$ Grid Planner         & 0.804 \\
                & $40cm$ Grid Planner         & 0.594 \\
                & Gradient Planner            & \textbf{0.998} \\\hline
\multirow{4}{*}{ReplicaCAD~\cite{szot2021habitat}}  & $10cm$ Grid Planner         & 0.974 \\
                & $20cm$ Grid Planner         & 0.913 \\
                & $30cm$ Grid Planner         & 0.606 \\
                & $40cm$ Grid Planner         & 0.523 \\
                & Gradient Planner            & \textbf{1.000} \\
\bottomrule
\end{tabular}
\end{table}

\subsection{Navigating to a Target Point}

In Table \ref{tab:path_lengths}, we compare the planners derived from our scene representation with our baseline planner, which uses the occupancy grid constructed from the raw point clouds. Since the latter planner doesn't have an explicit notion of semantics, we instead choose 100 random starting and ending point pairs for each of our scenes and compute the length of the paths between each point pair generated by each planner. We normalize by the shortest path length and average over all the point pairs in a given scene to generate our final metric.

\subsection{Navigating to a Semantic Location}

In Table \ref{tab:semantics}, we compare how well each planner does at navigating to the goal semantic location, by using the CLIP correspondence score between the linguistic embedding and the CLIP embedding for the spatial coordinate. While the latter is derived from the model, thus making this comparison somewhat self-referential, the purpose of this evaluation is primarily to compare the performance of each planner on the same fixed map, rather than evaluate the quality of the map itself. We find that the gradient planner is able to take advantage of the gradient information provided by the semantic embedding to significantly improve the CLIP embedding of the final location. We compare the locations empirically and find that the CLIP scores correspond well to the linguistic targets.

%% file: 06conclusion.tex
\model{} allows us to build an end-to-end differentiable motion planning system for autonomous robots based on an implicit scene memory which captures both affordances and semantic information. We see that our proposed gradient-based planner can use the \model{} representation to both better optimize arriving at semantic goals, and to generate better, shorter paths.

One potential direction for future research is to explore more complex planning strategies by adding additional objectives to the gradient planner. While our approach is already very flexible, there may be situations where more complex planning strategies are necessary to navigate through complex environments. For example, it may be beneficial to optimize for smoother paths, as in~\cite{ratliff2009chomp}. By incorporating additional objectives into the planner, we can potentially improve the performance and efficiency of the navigation system in these situations.

Another important direction for future work is to construct the representation while the environment is being explored. In this paper, we focused on building plans on representations that are already fully constructed. If we could construct \model{} while the environment is being explored, this might even enable language-driven exploration.